\documentclass[letterpaper, 10 pt, journal]{IEEEtran}
\IEEEoverridecommandlockouts                          
\usepackage{graphics} 
\usepackage{epsfig}
\usepackage{times}
\usepackage{amsmath}
\usepackage{amssymb}
\usepackage{threeparttable}
\usepackage{booktabs}
\usepackage{soul}
\usepackage{subfigure}
\usepackage{xcolor}
\usepackage[hidelinks]{hyperref}
\usepackage{multirow}

\usepackage{cuted}
\usepackage{capt-of}

\title{\LARGE \bf
NeurRAFT: \underline{R}obot Motion Planning via \underline{A}nchor-Level \underline{F}low Matching with Clearance-Aware Preference \underline{T}uning
}

\author{Sibo Tian$^{1}$, Chang Liu$^{1}$, Minghui Zheng$^{1,*}$, and Xiao Liang$^{2,*}$
\thanks{This work was supported by the USA National Science Foundation under Grant No. 2422826 and No. 2527316. Portions of this research were conducted with the advanced computing resources provided by Texas A\&M High Performance Research Computing.}
\thanks{$^{1}$ Sibo Tian, Chang Liu, and Minghui Zheng are with the J. Mike Walker '66 Department of Mechanical Engineering, Texas A\&M University, College Station, TX 77843, USA. {\tt\small Emails: {sibotian@tamu.edu, changliu.chris@tamu.edu, mhzheng@tamu.edu.}}}
\thanks{$^{2}$ Xiao Liang is with the Zachry Department of Civil and Environmental Engineering, Texas A\&M University, College Station, TX 77843, USA. {\tt\small Email: xliang@tamu.edu.}}
\thanks{$^{*}$ Corresponding Authors.}
}

\begin{document}

\maketitle
\thispagestyle{empty}
\pagestyle{empty}

\begin{strip}
\vspace{-1.3in}
\centering
\includegraphics[width=\textwidth]{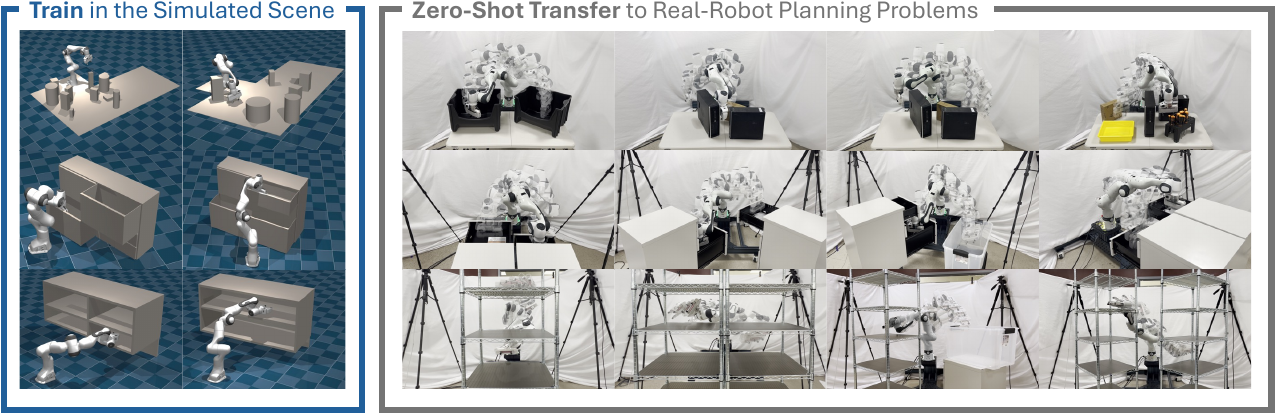}
\captionof{figure}{We present NeurRAFT, a generative motion planner trained entirely in simulation that transfers zero-shot to a physical Franka manipulator. \textit{Left:} training on the M$\pi$Nets benchmark \cite{fishman2023motion}, covering tabletop, cubby, and dresser scenes. \textit{Right:} the same model plans directly from depth observations in real-world settings with physical obstacles unseen during training. Each panel overlays the executed trajectory as a sequence of robot configurations.}
\label{fig:overview}
\end{strip}

\begin{abstract}

Recent end-to-end neural motion planners generate trajectories from raw sensor observations, avoiding the privileged geometric models required by classical planners. However, collision-free planning in cluttered environments remains challenging. We present NeurRAFT, a generative planning framework based on anchor-level flow matching and clearance-aware preference tuning. Unlike prior neural planners that model dense waypoint sequences and spend capacity on redundant local details and smoothness, NeurRAFT operates on compact anchor waypoints. We train the planner using a Jacobian-weighted loss that accounts for the task-space impact of each anchor. At inference, the anchors are generated in two integration steps, followed by cubic-spline interpolation to recover a smooth, full-resolution trajectory. Since imitation learning from positive demonstrations cannot distinguish collision-free from near-collision trajectories, collision-prone behaviors persist at test time. Rather than relying on post-hoc corrections, we directly reshape the pretrained planner's distribution toward safer solutions without augmenting inference. Specifically, Direct Preference Optimization shifts probability mass toward trajectories with larger obstacle clearance, with the resulting improvement directly absorbed into the planner parameters. Experiments show substantial improvements over state-of-the-art planners, while real-world experiments demonstrate zero-shot transfer to a Franka robot under noisy and partially occluded depth observations. Video results available at \textcolor{blue}{\url{https://neurraft.github.io/}}.

\end{abstract}

\begin{IEEEkeywords}
Neural Motion Planning, Flow Matching, Trajectory Parameterization, Direct Preference Optimization.
\end{IEEEkeywords}

\section{Introduction}

Motion planning for robot manipulators in cluttered, unstructured environments remains a longstanding challenge, as feasible motions must be generated in a high-dimensional configuration space that is only partially observable \cite{soleymanzadeh2026towards}. This is particularly critical in industrial applications such as robotic remanufacturing, where manipulators work over cluttered workbenches surrounded by fixtures, tools, and storage bins, and must generate collision-free motions from partial observations of a scene that changes as parts are removed or reassembled \cite{tian2026redefining}. Extensive research over the past decades has resulted in numerous well-studied motion planning algorithms. Classical approaches broadly fall into two families: sampling-based planners and optimization-based planners. Sampling-based planners, such as PRM \cite{kavraki2002probabilistic} and the RRT family \cite{lavalle1998rapidly, karaman2011sampling, kuffner2000rrt}, explore the configuration space by randomly sampling collision-free states and connecting them into a graph or tree, offering probabilistic completeness and asymptotic optimality. Optimization-based planners, including CHOMP \cite{ratliff2009chomp} and TrajOpt \cite{schulman2014motion}, instead formulate motion planning as a trajectory optimization problem, refining an initial trajectory to minimize an objective that penalizes collisions while encouraging smoothness. Despite their maturity, these classical methods rely on an explicit geometric model of the environment for collision checking, which is computationally expensive to query and difficult to reconstruct accurately from raw sensor data. Furthermore, sampling-based planners may require many iterations to find a feasible path in cluttered environments, whereas optimization-based planners are sensitive to initialization and prone to becoming trapped in local minima.

To overcome these limitations, recent research has increasingly turned to neural motion planners. Early learning-based approaches augment, rather than replace, classical planners. Instead of sampling the configuration space uniformly, several methods learn where to sample, biasing the sampling process toward regions likely to contain a feasible solution and thereby accelerating the underlying search \cite{ichter2018learning, wang2020neural}. Later neural planners, such as MPNet \cite{qureshi2020motion} and M$\pi$Net \cite{fishman2023motion}, learn to predict the next waypoint directly from the current state and a representation of the scene, amortizing the cost of search into neural network inference. While such informed prediction substantially accelerates planning, these methods still require multi-step rollout to generate a complete trajectory, and prediction errors can accumulate over the planning horizon. Moreover, these methods do not explicitly model the distribution over feasible trajectories. Even with stochasticity injected via dropout at inference \cite{qureshi2020motion, soleymanzadeh2025simpnet}, the diversity of generated paths is limited, whereas multiple topologically distinct trajectories may all be feasible in a cluttered environment.

To explicitly model the multimodality of robot motion, another line of neural motion planners casts motion planning as conditional generation. Diffusion-based planners \cite{janner2022planning, carvalho2025motion, saha2024edmp} have shown strong performance by generating entire trajectories at once and naturally representing diverse solution modes. However, diffusion models typically require many iterative denoising steps, making inference slow. More recent works adopt flow matching, which learns a velocity field along near-straight probability paths between the source and target distributions, enabling generation in far fewer integration steps \cite{tian2025warm}. Despite this progress, two limitations persist across existing generative planners. First, regardless of the underlying generative model, trajectories are usually represented as dense, full-length waypoint sequences. Such sequences are inherently redundant, as a smooth trajectory can be effectively characterized by a small number of control points, and modeling the remaining waypoints forces the network to spend capacity on local smoothness that interpolation provides for free. Prior efforts have primarily improved generative planning efficiency through the sampling process, while trajectory representations have received comparatively less attention. Second, previous methods are trained exclusively on successful demonstrations and are never exposed to undesirable trajectories. Consequently, they cannot distinguish collision-free motions from trajectories that graze or penetrate obstacles, leaving a residual collision rate that supervised training alone cannot eliminate.

To fill in the gap, in this work we present NeurRAFT, a generative robot motion planning framework that rethinks both the trajectory representation and the training signal. To eliminate representational redundancy, NeurRAFT plans at the level of anchor waypoints. Conditioned on a point-cloud observation of the scene, our method generates a compact sequence of anchors via flow matching in as few as two integration steps, and recovers a smooth, full-resolution trajectory by fitting a cubic spline that passes through the generated anchors. Because the manipulator's kinematics amplify some joint errors more than others, an isotropic loss in joint space treats all joints as equally important, whereas the same angular deviation at different joints can displace the end effector by very different amounts. In our anchor-based representation, each generated anchor corresponds to an actual configuration on the executed trajectory, unlike B-spline control points \cite{carvalho2025motion}, which shape the curve without necessarily lying on it. We therefore weight the velocity-regression error at each anchor using the Jacobian evaluated at that configuration, penalizing errors according to the end-effector displacement they induce.

Beyond the representation and loss design, we further address the limitation of supervised imitation learning discussed above. Existing approaches often improve planning success by post-hoc correcting failed or unsafe trajectories through additional inference-time procedures, such as refinement \cite{sharma2025cascaded}, guidance \cite{carvalho2025motion}, or trajectory optimization \cite{tian2025warm, huang2025diffusionseeder}. Rather than correcting undesirable trajectories after generation, we directly reshape the planner's output distribution at the source. We argue that the pretrained planner can already capture feasible solutions within its learned trajectory distribution, and the key challenge is therefore the imperfect allocation of probability mass within this distribution. We address this by introducing clearance-aware preference tuning, where trajectories sampled from the pretrained planner are ranked by signed obstacle clearance and used to construct preference pairs. We then fine-tune the planner with direct preference optimization (DPO) \cite{rafailov2023direct} to shift probability mass toward higher-clearance solutions. Since the preference data are generated on the training dataset, this stage requires neither additional demonstrations nor human annotations. The resulting improvement is directly absorbed into the planner parameters, leaving the inference procedure unchanged. An overview can be seen in Fig.~\ref{fig:overview}. Overall, our contributions are summarized as follows:

\begin{itemize}
    \item We propose NeurRAFT, a generative robot motion planner that replaces dense trajectory modeling with compact on-trajectory anchors. The planner is trained with a Jacobian-weighted flow matching loss to account for task-space impact, while inference generates the anchors in only two integration steps and recovers the full trajectory through cubic-spline interpolation.
    
    \item We introduce a preference alignment strategy that improves planning success by reshaping the planner's trajectory distribution toward safer solutions, rather than relying on post-hoc inference-time corrections. Ranking the planner's own samples by obstacle clearance provides the negative supervision missing from imitation learning. To our knowledge, this is the first use of self-generated, clearance-based preference alignment for collision avoidance in neural motion planning.
    
    \item We conduct extensive experiments on the M$\pi$Nets benchmark, demonstrating substantial improvements in planning success rate over state-of-the-art planners. Real-world experiments further show zero-shot transfer from simulation to a physical Franka robot under noisy and partially occluded depth observations.
\end{itemize}

\section{Related Work}

\subsection{Neural Motion Planning}

Machine learning has been integrated into motion planning at different levels. One line of work retains the classical search structure but learns informed sampling distributions. Ichter et al. \cite{ichter2018learning} train a conditional variational autoencoder on successful plans to bias sampling toward promising regions, while Neural RRT* \cite{wang2020neural} predicts a sampling probability heatmap from the workspace layout. These methods accelerate search but still require an explicit collision checker in the loop. Another line of work directly predicts robot motions rather than guiding a classical planner. MPNet \cite{qureshi2020motion} encodes the workspace point cloud into a latent representation and autoregressively predicts the next robot configuration, relying on collision checking with privileged geometric models to validate local connections during recursive path generation. In contrast, M$\pi$Net \cite{fishman2023motion} learns an end-to-end motion policy from a large corpus of expert trajectories, iteratively predicting configuration displacements to generate collision-free trajectories without relying on privileged knowledge at inference. Follow-up work improves the scene encoder and policy head \cite{soleymanzadeh2025simpnet}. However, these autoregressive planners emit one waypoint per forward pass, requiring tens of sequential rollouts during trajectory generation, where prediction errors can compound. Their policies also capture limited multimodality even with inference-time dropout \cite{qureshi2020motion}.


Casting motion planning as conditional generation provides a natural way to model multimodal distributions over feasible trajectories, where multiple topologically distinct paths may solve the same planning problem. Diffuser \cite{janner2022planning} pioneered the joint denoising of entire state-action sequences, avoiding the error accumulation of autoregressive rollouts and enabling test-time objective guidance. Motion Planning Diffusion \cite{carvalho2025motion} first applied trajectory-level diffusion to manipulator motion planning, steering denoising with gradients of planning costs, including a collision cost that requires a precomputed signed distance field (SDF) at inference. EDMP \cite{saha2024edmp} extends this scheme with an ensemble of collision-cost guides. Cascaded Diffusion \cite{sharma2025cascaded} instead employs a coarse-to-fine hierarchy of diffusion models, where a coarse model generates global sub-goals that condition a lower-level diffusion model. Residual collisions are then addressed by re-running the local model on violating segments, requiring explicit collision detection during inference. Thus, existing generative planners often rely on some form of privileged collision information or additional inference-time procedures for collision avoidance, while also inheriting the many-step sampling cost of diffusion. Flow matching \cite{lipman2023flow} offers a more efficient alternative by learning a velocity field along near-straight probability paths, enabling high-quality generation in only a few integration steps, and has recently been applied to robot motion planning for warm-starting trajectory optimization \cite{tian2025warm}.


Moreover, beyond the best-of-$N$ strategy \cite{carvalho2025motion, saha2024edmp, dalal2024neural} that comes naturally with stochastic methods, to improve planning success, state-of-the-art methods typically rely on post-hoc mechanisms, including hierarchical planning \cite{sharma2025cascaded, chen2024simple}, test-time guidance \cite{carvalho2025motion, saha2024edmp}, trajectory refinement \cite{sharma2025cascaded}, and subsequent trajectory optimization \cite{tian2025warm, huang2025diffusionseeder}. These approaches improve planning performance by augmenting the inference pipeline with additional generation, refinement, or optimization procedures. In contrast, we take a different perspective: rather than correcting undesirable trajectories after their initial generation, we reshape the planner's learned distribution itself toward safer solutions. This reduces the probability of generating low-clearance trajectories in the first place, increasing the likelihood of sampling high-clearance, feasible solutions under a fixed sampling budget without introducing additional inference-time computation.

\subsection{Direct Preference Optimization for Preference Alignment}
Direct preference optimization (DPO) \cite{rafailov2023direct} converts ranked output pairs into a contrastive likelihood objective, removing the explicit reward model of reinforcement learning from human feedback (RLHF). Originally developed for language models and later adapted to image diffusion models \cite{wallace2024diffusion}, DPO has recently entered embodied domains. Yuan et al. \cite{yuan2025preference} fine-tune a quadrupedal locomotion diffusion controller through online interaction, using a weak preference labeling scheme that requires neither ground-truth rewards nor human annotations. FlowPRO \cite{wu2026flowpro} derives a proximalized preference objective for the flow matching action head of vision-language-action models, constructing preference pairs from teleoperated human interventions on a real robot. Across these works, preferences are judged by task performance via online rollouts or human intervention, and constructing them requires either additional online interaction in simulation or teleoperation on a real robot. In contrast, NeurRAFT derives preferences from a purely offline geometric quantity, namely the signed obstacle clearance computed on the model's own samples, requiring neither online interaction nor human involvement, and aimed specifically at the residual collision rate that imitation-only training can hardly eliminate. To the best of our knowledge, NeurRAFT is the first neural motion planner for robot manipulators to use self-generated, clearance-based preference alignment for collision avoidance, without requiring additional demonstrations, human annotations, or online interaction.

\section{Methodology}

\subsection{Problem Formulation}

\begin{figure*}
    \begin{center}
        \includegraphics[width=2.0\columnwidth]{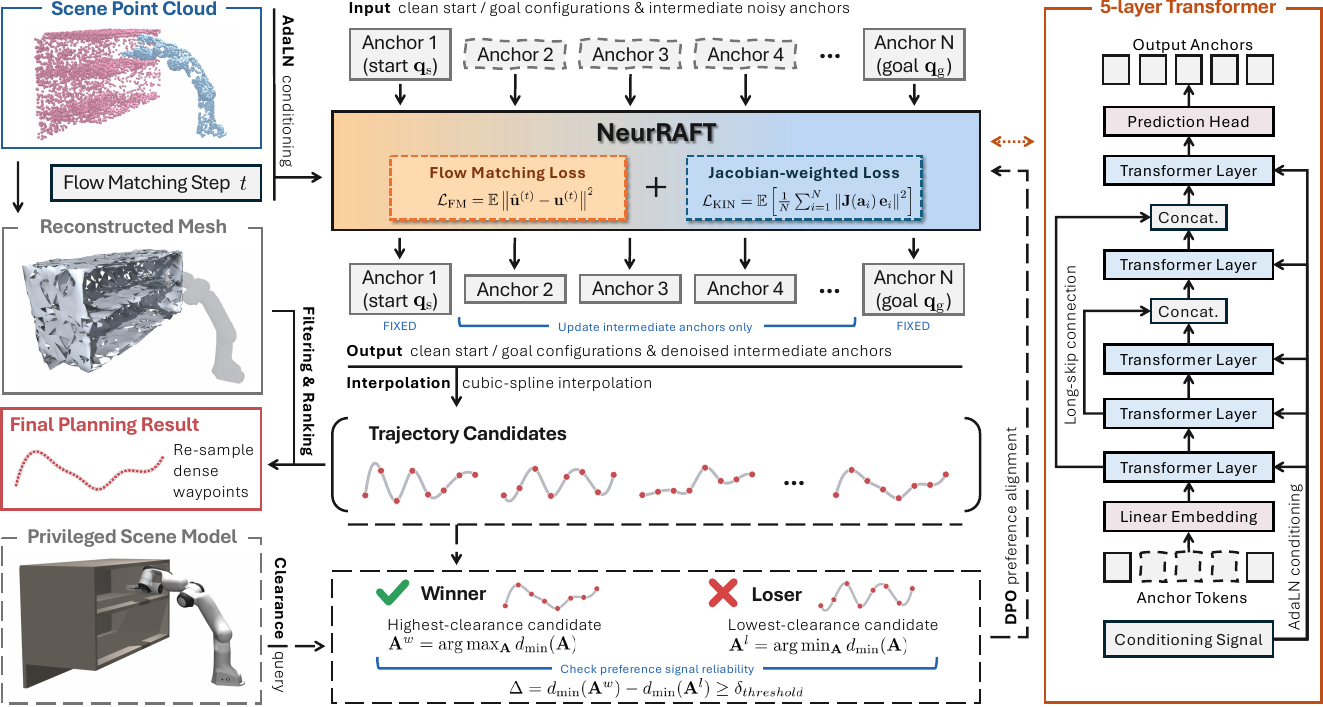}
    \vspace{-0.1in}
    \caption{Overview of NeurRAFT. Given a scene point cloud and clean start/goal configurations, NeurRAFT generates a compact set of trajectory anchors via flow matching, with the intermediate anchors progressively denoised while the boundary configurations remain fixed. A cubic spline then interpolates the anchors to recover smooth trajectory candidates, which are filtered and ranked against a mesh reconstructed from the observed point cloud, and the selected candidate is resampled into dense waypoints as final planning solution. During preference alignment (dashed process), the trajectories are evaluated using a privileged scene model to obtain signed obstacle clearance, and preference pairs are constructed to fine-tune NeurRAFT toward higher-clearance trajectories. This privileged information is used only during training and is not required at inference, where NeurRAFT plans directly from the scene point cloud.}
    \label{fig:framework}
    \end{center}
    \vspace{-0.2in}
\end{figure*}

We consider the problem of collision-free motion planning for a robot manipulator with $d$ degrees of freedom (DoF), operating in a cluttered environment observed only through sensor measurements. Let $\mathcal{Q} \subseteq \mathbb{R}^{d}$ denote the configuration space of the manipulator. The planner receives a point cloud observation $\mathbf{O} \in \mathbb{R}^{P \times 3}$ of the scene, a start configuration $\mathbf{q}_{\mathrm{s}} \in \mathcal{Q}$, and a goal end-effector pose $\mathbf{T}_{\mathrm{g}} \in SE(3)$, and must produce a trajectory that starts from $\mathbf{q}_{\mathrm{s}}$ and reaches $\mathbf{T}_{\mathrm{g}}$ while avoiding collisions with the environment. We use $\mathbf{q}_{\mathrm{g}}$ to denote a goal configuration that reaches $\mathbf{T}_{\mathrm{g}}$, and for redundant manipulators, multiple possible goal configurations may exist. Crucially, no privileged geometric model of the scene is available at inference time, and all information about obstacles must be extracted from the raw point cloud.

Prior generative planners typically represent a trajectory as a dense sequence of $H$ waypoints
\begin{equation}
    \boldsymbol{\xi} = \left[ \mathbf{q}_1, \mathbf{q}_2, \ldots, \mathbf{q}_H \right], \quad \mathbf{q}_t \in \mathcal{Q},
\end{equation}
where $H$ is large enough for consecutive waypoints to be densely spaced along the path. Instead, we parameterize the trajectory by a compact sequence of $N$ anchor waypoints
\begin{equation}
    \mathbf{A} = \left[ \mathbf{a}_1, \mathbf{a}_2, \ldots, \mathbf{a}_N \right], \quad \mathbf{a}_i \in \mathcal{Q}, \quad N \ll H,
\end{equation}
where the first and last anchors are constrained to the start and goal configurations, i.e., $\mathbf{a}_1 = \mathbf{q}_{\mathrm{s}}$ and $\mathbf{a}_N = \mathbf{q}_{\mathrm{g}}$. Each anchor is a configuration that the trajectory passes through, and the full trajectory is recovered by a deterministic reconstruction operator $\mathcal{S}$ that fits a cubic spline through the anchors:
\begin{equation}
    \tau = \mathcal{S}(\mathbf{A}), \qquad
    \tau(0) = \mathbf{q}_{\mathrm{s}}, \quad
    \tau(1) = \mathbf{q}_{\mathrm{g}},
\end{equation}
where $\tau : [0,1] \rightarrow \mathcal{Q}$ denotes the continuous-time trajectory. For trajectory evaluation and downstream processing, we discretize the recovered trajectory into $H$ uniformly sampled waypoints denoted as $\tau_{\mathbf{A}}=\{\tau(t_h)\}_{h=1}^{H}$, where $t_h \in [0,1]$ denotes the normalized parameter of the $h$-th waypoint. Since $\mathcal{S}$ interpolates the anchors exactly and cubic splines are twice continuously differentiable, smoothness of the recovered trajectory and exact satisfaction of the boundary conditions are obtained by construction. Such parameterization reduces the dimensionality of the generation problem from $H \times d$ for a dense trajectory of $H$ waypoints to $N \times d$ with $N \ll H$, relieving the network of modeling local smoothness that interpolation provides for free. Moreover, since every anchor is a configuration that the manipulator actually passes through, the representation remains physically interpretable, with each generated variable corresponding to a reachable robot state rather than an abstract curve parameter.

In this work, we treat robot motion planning as a conditional generation problem. Given the context $\mathbf{c} = (\mathbf{O}, \mathbf{q}_{\mathrm{s}}, \mathbf{T}_{\mathrm{g}})$, we learn the conditional distribution $p(\mathbf{A} | \mathbf{c})$ over anchor sequences whose induced trajectories $\mathcal{S}(\mathbf{A})$ are collision-free and satisfy the start and goal constraints. Modeling a distribution rather than regressing a single solution is essential for this problem. In cluttered scenes, multiple topologically distinct trajectories may all be feasible, such as passing an obstacle on either side. A deterministic predictor conditioned on the same context would be forced to average over these modes, producing solutions that interpolate between valid paths and potentially cut through obstacles. A generative formulation instead preserves this multimodality. It allows the planner to sample diverse candidates from $p(\mathbf{A} | \mathbf{c})$, improving the chance of finding a feasible solution. An overview of our framework is illustrated in Fig.~\ref{fig:framework}.

\subsection{Anchor-Level Flow Matching}

We model the conditional distribution $p(\mathbf{A}| \mathbf{c})$ with flow matching over the anchor sequences $\mathbf{A} \in \mathbb{R}^{N \times d}$ defined above. From each dense expert trajectory, we subsample $N$ configurations at uniform arc-length spacing to form a data sample $\mathbf{A}^{(1)}$, and we fix the start and goal configurations as the first and last anchors. We then define a linear probability path from a Gaussian source sample $\mathbf{A}^{(0)} \sim \mathcal{N}(\mathbf{0}, \mathbf{I})$ to a data sample $\mathbf{A}^{(1)}$ drawn from the expert distribution, i.e.,
\begin{equation}
    \mathbf{A}^{(t)} = (1 - t)\,\mathbf{A}^{(0)} + t\,\mathbf{A}^{(1)},
    \qquad t \in [0, 1],
    \label{eq:interp}
\end{equation}
whose corresponding target velocity is the straight-line displacement

\begin{equation}
    \mathbf{u}^{(t)} = \mathbf{A}^{(1)} - \mathbf{A}^{(0)} = (\mathbf{A}^{(1)} - \mathbf{A}^{(t)})/(1 - t).
    \label{eq:velo_field}
\end{equation}
All anchors share a single flow time $t$, except the two endpoints. We exempt the endpoint anchors from the interpolation and hold them at $t_1 = t_N = 1$ throughout training and sampling, clamping their values to $\mathbf{q}_{\mathrm{s}}$ and $\mathbf{q}_{\mathrm{g}}$. This imposes the boundary conditions by inpainting. During training, the network always observes clean start and goal anchors, and we supervise its predicted velocities at all $N$ anchors, with the two boundary anchors carrying zero target velocity. At inference we reset the endpoint anchors to $\mathbf{q}_{\mathrm{s}}$ and $\mathbf{q}_{\mathrm{g}}$ after each integration step, so the boundary conditions hold exactly.

The network $f_\theta$ adopts the $\mathbf{A}^{(1)}$-prediction
parameterization \cite{li2025back}. Given the partially noised anchor sequence $\mathbf{A}^{(t)}$, the planning problem context $\mathbf{c}$, and the per-anchor times $t$, the network generates the clean anchor sequence as the model output, i.e.,

\begin{equation}
\hat{\mathbf{A}}^{(1)} = f_\theta(\mathbf{A}^{(t)}, \mathbf{c}, t),
\end{equation}
from which the velocity estimate can be recovered by
\begin{equation}
\hat{\mathbf{u}}^{(t)} = (\hat{\mathbf{A}}^{(1)} - \mathbf{A}^{(t)})/(1-t).
\end{equation}
During implementation, we clip the denominator to a small positive value to avoid numerical instability. Then flow matching objective minimize the squared error between this velocity and the target,
\begin{equation}
    \mathcal{L}_{\mathrm{FM}}
    = \mathbb{E}_{\mathbf{c},\, \mathbf{A}^{(1)},\, \mathbf{A}^{(0)},\, t}
      \left\| \hat{\mathbf{u}}^{(t)} - \mathbf{u}^{(t)} \right\|^2 .
    \label{eq:fm}
\end{equation}
The flow matching loss weights every anchor and joint equally. Specifically, it is a uniform squared error in configuration space, such that a unit error on a base joint contributes the same loss as a unit error on a wrist joint, even though the two can induce substantially different end-effector displacements. The same joint error can also have different task-space effects across anchors, depending on the robot configuration. Since the anchors represent valid robot configurations, we can additionally weight the regression error according to its task-space effect. Let $\mathbf{J}(\mathbf{a}_i) \in \mathbb{R}^{6 \times d}$ denote the analytical manipulator Jacobian evaluated at the ground-truth anchor. Defining the per-anchor velocity error $\mathbf{e}_i = \hat{\mathbf{u}}^{(t)}_i - \mathbf{u}^{(t)}_i$, we add a kinematics-aware term
\begin{equation}
    \mathcal{L}_{\mathrm{KIN}}
    = \mathbb{E} \left[ \tfrac{1}{N} \textstyle\sum_{i=1}^{N}
      \left\| \mathbf{J}(\mathbf{a}_i)\, \mathbf{e}_i \right\|^2 \right],
    \label{eq:jac}
\end{equation}
which penalizes configuration-space errors in proportion to the end-effector motion they induce. The final training objective combines the standard flow matching loss with the proposed kinematics-aware term, i.e., $\mathcal{L} = \mathcal{L}_{\mathrm{FM}} + \mathcal{L}_{\mathrm{KIN}}$.

We use a simple Transformer with long-skip connections between shallow and deep layers as our backbone. The scene point cloud, comprising labeled robot and obstacle points, is encoded by a PointNet++ encoder \cite{qi2017pointnet++} into a global context feature, which is combined with the per-anchor flow time embeddings to form anchor-specific conditioning features. These conditioning features are injected into every Transformer layer through adaptive layer normalization (AdaLN) \cite{peebles2023scalable}. Specifically, each attention and feed-forward block is modulated by learned shift, scale, and gate vectors that are computed independently for each anchor from its corresponding conditioning feature.

At inference, we integrate the learned flow from $t = 0$ to $t = 1$ using an explicit Euler scheme, while enforcing the endpoint constraints after each integration step. To exploit the redundancy of the robot manipulator and promote global diversity, we generate multiple candidate trajectories corresponding to different goal configurations that achieve the same goal end-effector pose $\mathbf{T}_{\mathrm{g}}$. Given the target $\mathbf{T}_{\mathrm{g}}$, we first solve inverse kinematics to obtain multiple feasible goal configurations $\mathbf{q}_{\mathrm{g}}$, and generate a batch of candidate trajectories in parallel, each conditioned on a different goal configuration. We then use the observed point cloud to reconstruct the scene mesh and perform collision checking to filter out candidates that may result in collisions. Importantly, collision checking at this stage does not rely on any privileged geometric model, and it is performed solely using the observed point cloud. Among the candidates that pass the collision filter, we select the one with the shortest end-effector path length. If none of the candidates is collision-free, we instead select the candidate with the minimum collision depth as the final solution.

\subsection{Clearance-Aware Preference Tuning}

Flow matching learns to imitate the expert trajectory distribution from successful demonstrations, but imitation learning provides only positive examples and does not explicitly teach the planner what constitutes an undesirable trajectory. Consequently, although the learned distribution captures the behavior of the experts, it provides no explicit penalty for trajectories that graze or slightly penetrate obstacles. As a result, some collision-prone behaviors can persist at test time, leaving the imitation-only planner unable to reliably eliminate such collisions. Recent approaches therefore improve the success rate of neural motion planners by introducing additional components or procedures at inference time, including hierarchical planning \cite{sharma2025cascaded, chen2024simple}, classifier guidance during sampling \cite{carvalho2025motion, saha2024edmp}, collision-aware refinement of generated trajectories \cite{sharma2025cascaded}, or a separate trajectory-optimization stage initialized from the planner output \cite{tian2025warm, huang2025diffusionseeder}. While effective, these approaches require additional computation or planning procedures beyond the pretrained neural planner during inference. We instead take a different perspective: the pretrained planner already captures feasible solutions within its learned trajectory distribution, such that a sufficiently diverse set of samples can contain at least one feasible trajectory. This suggests that the fundamental limitation is not the absence of feasible solutions, but the imperfect allocation of probability mass within the learned distribution. Motivated by this, we directly reshape the pretrained distribution toward trajectories with larger geometric clearance through an offline preference alignment stage. Specifically, we rank candidate trajectories according to their signed obstacle clearance and fine-tune the planner to assign greater preference to higher-clearance trajectories. The resulting improvement is therefore absorbed directly into the planner weights, allowing the original sampling procedure to remain unchanged at inference time.

Specifically, for each training context $\mathbf{c}$, we draw multiple candidate anchor sequences from the pretrained planner and reconstruct each set into a dense waypoint trajectory. We denote the resulting trajectory by $\tau_{\mathbf{A}}(t)$. We then evaluate each reconstructed trajectory using a single signed-clearance scalar
\begin{equation}
d_{\min}(\mathbf{A}) =
\min_{t \in \{1,\ldots,H\}}
\mathrm{sd}\!\left(
\mathcal{R}(\tau_{\mathbf{A}}(t)), \mathcal{E}
\right),
\label{eq:dmin}
\end{equation}
where $\mathrm{sd}(\cdot,\cdot)$ denotes the signed distance between the robot geometry $\mathcal{R}$ at configuration $\tau_{\mathbf{A}}(t)$ and the environment $\mathcal{E}$. Positive values indicate collision-free clearance, while negative values indicate penetration, with larger negative values corresponding to deeper penetration. Each context yields one preference pair, with the winner
$\mathbf{A}^{w} = \arg\max_{\mathbf{A}} d_{\min}(\mathbf{A})$
and the loser
$\mathbf{A}^{l} = \arg\min_{\mathbf{A}} d_{\min}(\mathbf{A})$.
This construction provides informative comparisons across three regimes: when only some candidates are collision-free, it separates feasible from infeasible trajectories; when all candidates are in collision, it favors the trajectory with the shallowest penetration; and when all candidates are collision-free, it favors the one with the largest clearance margin. Preference pairs with a clearance gap
$\Delta = d_{\min}(\mathbf{A}^{w}) - d_{\min}(\mathbf{A}^{l})$
below a predefined threshold are discarded because they provide insufficiently reliable preference signals.

We leverage Direct Preference Optimization (DPO) for preference alignment. DPO formulates preference learning through an implicit reward induced by the policy \cite{rafailov2023direct}. However, trajectory log-likelihoods are intractable for our flow-based planner. Following the formulation in \cite{wu2026flowpro}, we use the per-sample flow matching loss as a tractable surrogate for the negative log-likelihood and define the corresponding implicit reward as
\begin{equation}
    r_\theta(\mathbf{A} \mid \mathbf{c}) =
    -\mathcal{L}_{\mathrm{FM}}
    \left(\theta;\mathbf{A},\mathbf{c}\right),
    \label{eq:reward}
\end{equation}
where the loss is evaluated only on the interior anchors, since the pinned endpoints are identical across candidates and therefore provide no preference signal. Let $f_{\mathrm{ref}}$ denote the frozen pretrained planner and $f_\theta$ the trainable policy initialized from it. The preference objective is then
\begin{equation}
    \mathcal{L}_{\mathrm{DPO}}
    = -\,\mathbb{E}\!\left[
    \log \sigma \left(
    \beta \left[
      \left(r_\theta^{w}-r_\theta^{l}\right)
      -
      \left(r_{\mathrm{ref}}^{w}-r_{\mathrm{ref}}^{l}\right)
    \right]
    \right)
    \right],
    \label{eq:dpo}
\end{equation}
where $\beta$ is the preference strength that scales the preference term. To reduce the variance of the single-sample surrogate, we share the flow time and source noise across the four reward evaluations within each preference pair and average the rewards over a small number of Monte Carlo draws. Finally, we retain the standard flow matching objective on expert demonstrations to prevent distribution shift, and optimize
\begin{equation}
    \mathcal{L}
    = \mathcal{L}_{\mathrm{DPO}}
    + \mathcal{L}_{\mathrm{FM}},
    \label{eq:total}
\end{equation}
where the flow matching term preserves the expert behavior while the preference term reshapes the learned distribution toward trajectories with larger geometric clearance. Importantly, this alignment is absorbed directly into the planner parameters, leaving the sampling procedure unchanged and requiring no post-hoc process such as additional refinement module or optimization step at inference. All privileged information used to construct these preferences, including ground-truth obstacle geometry available in the physics simulator, is used exclusively during training. The planner receives the same inputs at inference as those used by the pretrained model.

\section{Experiments}

\begin{table*}[htbp]
\caption{Success rates (\%) on the Global-Solvable, Hybrid-Solvable, and Both-Solvable test splits of the M$\pi$Nets benchmark}
\centering
\resizebox{2.0\columnwidth}{!}{%
\begin{threeparttable}
\begin{tabular}{lllcc|cc|cc}
\toprule
\multirow{2}{*}{\raisebox{-1.0ex}{\textbf{Category}}}
&
\multirow{2}{*}{\raisebox{-1.0ex}{\textbf{Model}}}
&
\multirow{2}{*}{\raisebox{-1.0ex}{\textbf{Planning Strategy}}}
&
\multicolumn{2}{c|}{\textbf{Global-Solvable Split}}
&
\multicolumn{2}{c|}{\textbf{Hybrid-Solvable Split}}
&
\multicolumn{2}{c}{\textbf{Both-Solvable Split}}
\\
\cmidrule(lr){4-5}
\cmidrule(lr){6-7}
\cmidrule(lr){8-9}
&
&
&
\textbf{Glob. Expert} & \textbf{Hyb. Expert}
&
\textbf{Glob. Expert} & \textbf{Hyb. Expert}
&
\textbf{Glob. Expert} & \textbf{Hyb. Expert}
\\
\midrule
\multirow{5}{*}{\shortstack{\textit{Classical}\\\textit{Planner}}}
& Global Planner \cite{strub2020adaptively}
& Search, GT
& \multicolumn{2}{c|}{$\dagger$}
& \multicolumn{2}{c|}{78.44}
& \multicolumn{2}{c}{$\dagger$}
\\
& Hybrid Planner \cite{strub2020adaptively, van2022geometric}
& Search + Local control, GT
& \multicolumn{2}{c|}{50.22}
& \multicolumn{2}{c|}{$\dagger$}
& \multicolumn{2}{c}{$\dagger$}
\\
& CHOMP \cite{zucker2013chomp}
& Optimization, GT
& \multicolumn{2}{c|}{26.67}
& \multicolumn{2}{c|}{31.61}
& \multicolumn{2}{c}{32.20}
\\
& Geometric Fabrics \cite{van2022geometric}
& Reactive control, GT
& \multicolumn{2}{c|}{38.44}
& \multicolumn{2}{c|}{59.33}
& \multicolumn{2}{c}{60.06}
\\
& STORM \cite{bhardwaj2022storm}
& Sampling MPC, GT
& \multicolumn{2}{c|}{50.22}
& \multicolumn{2}{c|}{74.50}
& \multicolumn{2}{c}{76.00}
\\
\midrule
\multirow{7}{*}{\shortstack{\textit{Neural}\\\textit{Planner}}}
& MPNets \cite{qureshi2020motion}
& Single rollout + Collision check, GT
& -- & 41.33
& -- & 65.28
& -- & 67.67
\\
& M$\pi$Nets \cite{fishman2023motion}
& Single rollout until goal or timeout
& 75.06 & 75.78
& 80.39 & 95.33
& 82.78 & 95.06
\\
& EDMP \cite{saha2024edmp}
& Gen. + Cost guidance, GT
& 71.67 & 75.93
& 82.84 & 86.13
& 82.79 & 85.06
\\
& Neural MP \cite{dalal2024neural}
& Gen. + SDF selection
& 77.93 & 76.33
& 85.50 & 97.28
& 87.67 & 96.78
\\
& Cascaded Diffusion \cite{sharma2025cascaded}
& Gen. + Hierarchical + Refinement
& 85.13 & --
& -- & 98.00
& -- & --
\\
\cmidrule(lr){2-9}
& NeurRAFT - Pretraining Only
& Gen. + Mesh filtering
& 89.56 & 85.17
& 93.17 & 96.83
& 94.33 & 96.67
\\
& NeurRAFT
& Gen. + Mesh filtering
& \textbf{93.61} & \textbf{89.56}
& \textbf{97.89} & \textbf{98.22}
& \textbf{97.94} & \textbf{98.17}
\\
\bottomrule
\end{tabular}
\begin{tablenotes}
\item[*] Global-Solvable Split, Hybrid-Solvable Split, and Both-Solvable Split denote the three test splits of the benchmark. Glob. Expert and Hyb. Expert denote training on demonstrations from the Global and Hybrid planners, respectively. Classical planners require no training data, so a single value is reported across both training settings. Within the Classical Planner category, ``$\dagger$'' indicates a trivial success rate of 100\% by definition of the corresponding test split, whereas within the Neural Planner category, ``--'' indicates that the result is not reported in the original paper. Bold values indicate the best success rate within each column. Planning Strategy summarizes what each planner does at test time. GT indicates that an explicit ground-truth model of the scene is queried during planning. Gen. denotes the generative model, which generates multiple candidates and selects one as the final trajectory.
\end{tablenotes}
\label{tab:success_rate}
\end{threeparttable}
}
\vspace{-0.2in}
\end{table*}

\subsection{Experimental Setup}

\textbf{Datasets:} We evaluate NeurRAFT on the large-scale motion planning benchmark introduced by M$\pi$Nets \cite{fishman2023motion}, which considers a 7-DoF Franka robot manipulator operating in cluttered scenes across three categories: tabletop, cubby, and dresser. The benchmark provides two expert data-generation pipelines. The \emph{global} planner solves each problem directly in configuration space using a sampling-based optimal planner, AIT* \cite{strub2020adaptively}, whereas the \emph{hybrid} planner combines global end-effector planning via AIT* with a local geometric fabric controller \cite{van2022geometric}, producing smoother and more consistent demonstrations. The global expert dataset contains 6.54M expert trajectories generated across 773K environments, while the hybrid expert dataset contains 3.27M trajectories generated across 576K environments. Expert trajectories are discretized into $H=50$ dense waypoints in the original dataset. We train separate models on the two datasets to remain consistent with existing works and enable direct comparisons with prior methods. The benchmark further provides three held-out test sets: \emph{Global-Solvable}, \emph{Hybrid-Solvable}, and \emph{Both-Solvable}. The \emph{Global-Solvable} split consists of problems solvable by the global expert but not by the hybrid expert, while the \emph{Hybrid} split contains problems solvable by the hybrid expert but not by the global expert. The \emph{Both} split contains problems that can be successfully solved by both expert pipelines. Each split contains 600 problems from each of the three scene categories, yielding 1,800 planning problems per split.

\textbf{Evaluation Metrics:} Following \cite{fishman2023motion}, a trial is considered a \emph{success} if the executed trajectory reaches the target end-effector pose while remaining collision-free throughout its dense reconstruction. Success is verified in a physics simulator using the ground-truth scene geometry and a high-fidelity robot collision model, while also enforcing joint limits and self-collision constraints. We report the overall \emph{success rate} for each test split. To disentangle the quality of the learned generative model from the candidate-selection procedure, we additionally report the \emph{feasibility rate} for our method, defined as the fraction of all sampled candidate trajectories that are entirely collision-free. This metric measures the raw reliability of the learned trajectory distribution prior to any candidate filtering or selection in our framework.

To further characterize the effects of trajectory and goal diversity on planning performance, we evaluate our method under four sampling regimes: (1) one sample from a single goal configuration, (2) five samples from a single goal configuration, (3) one sample from each feasible inverse-kinematics (IK) goal, and (4) five samples from each feasible IK goal. For the single-goal regimes, we select the feasible IK solution that reaches the target end-effector pose and has the smallest joint-space distance to the initial configuration. These regimes progressively introduce trajectory-level and goal-level diversity, allowing us to disentangle the benefit of sampling multiple trajectories for a fixed goal from that of exploiting multiple feasible IK goals enabled by the manipulator's kinematic redundancy. For cases in which multiple candidate trajectories are generated, we apply the candidate-selection pipeline described in the previous section to rank and select the final trajectory for the success rate evaluation.

\textbf{Baselines:} We evaluate NeurRAFT against a broad set of classical and learning-based motion planners. Among the classical baselines, we include the global planner described in the previous section, which uses AIT* directly in configuration space \cite{strub2020adaptively}, as well as the hybrid planner combining task-space AIT* with a Geometric Fabrics local controller \cite{van2022geometric}. We also include CHOMP \cite{zucker2013chomp}, Geometric Fabrics \cite{van2022geometric}, and STORM \cite{bhardwaj2022storm} as our classical baselines.  We further consider recent neural motion planning methods, including MPNet \cite{qureshi2020motion}, M$\pi$Nets \cite{fishman2023motion}, EDMP \cite{saha2024edmp}, Neural MP \cite{dalal2024neural}, and Cascaded Diffusion \cite{sharma2025cascaded}, as our baselines.

\textbf{Implementation Details:} We use a 5-layer Transformer with long-skip connections linking the first and fifth layers and the second and fourth layers as the backbone of our method. The model has a latent dimension of 512, 8 attention heads, a feed-forward dimension of 1,024, and a dropout rate of 0.2. We downsample the obstacle point cloud to 4,096 points and sample 2,048 points from the robot in its start configuration, forming a segmented point cloud with 6,144 points in total. The scene point cloud is encoded by a PointNet++ encoder into a 2,048-dimensional conditioning feature. Flow matching pretraining is performed for 10 epochs with a batch size of 128 and a learning rate of $1\times10^{-4}$.

For the preference tuning stage, due to the large size of the original dataset, we first subsample the training problems using a fixed stride and perform inference with the pretrained model on the resulting subset. Preference pairs are constructed using a minimum clearance margin of 0.01 m, yielding 24,125 preference pairs for the global expert dataset and 6,583 preference pairs for the hybrid expert dataset. The policy is optimized with a batch size of 16 and a learning rate of $1\times10^{-5}$ for 1 epoch on the global expert dataset and 4 epochs on the hybrid expert dataset. Each surrogate reward is estimated by averaging over 4 Monte Carlo draws of the flow time and source noise. For each draw, the same flow time and noise are shared across the four evaluations of a preference pair, including the policy and reference models for both the winner and loser, allowing the stochasticity to largely cancel in the reward differences of Eq.~\eqref{eq:dpo}. The preference strength is set to $\beta=100$ for our model. We use five samples from each of the $K$ feasible IK goals as the default candidate generation pipeline, unless otherwise stated. All training and benchmark evaluations are conducted on a single NVIDIA A100 GPU. Inference time and Real-world experiments are evaluated on a single NVIDIA RTX 4080 GPU.

\subsection{Comparison with Baselines}

\begin{figure*}
    \begin{center}
        \includegraphics[width=2.0\columnwidth]{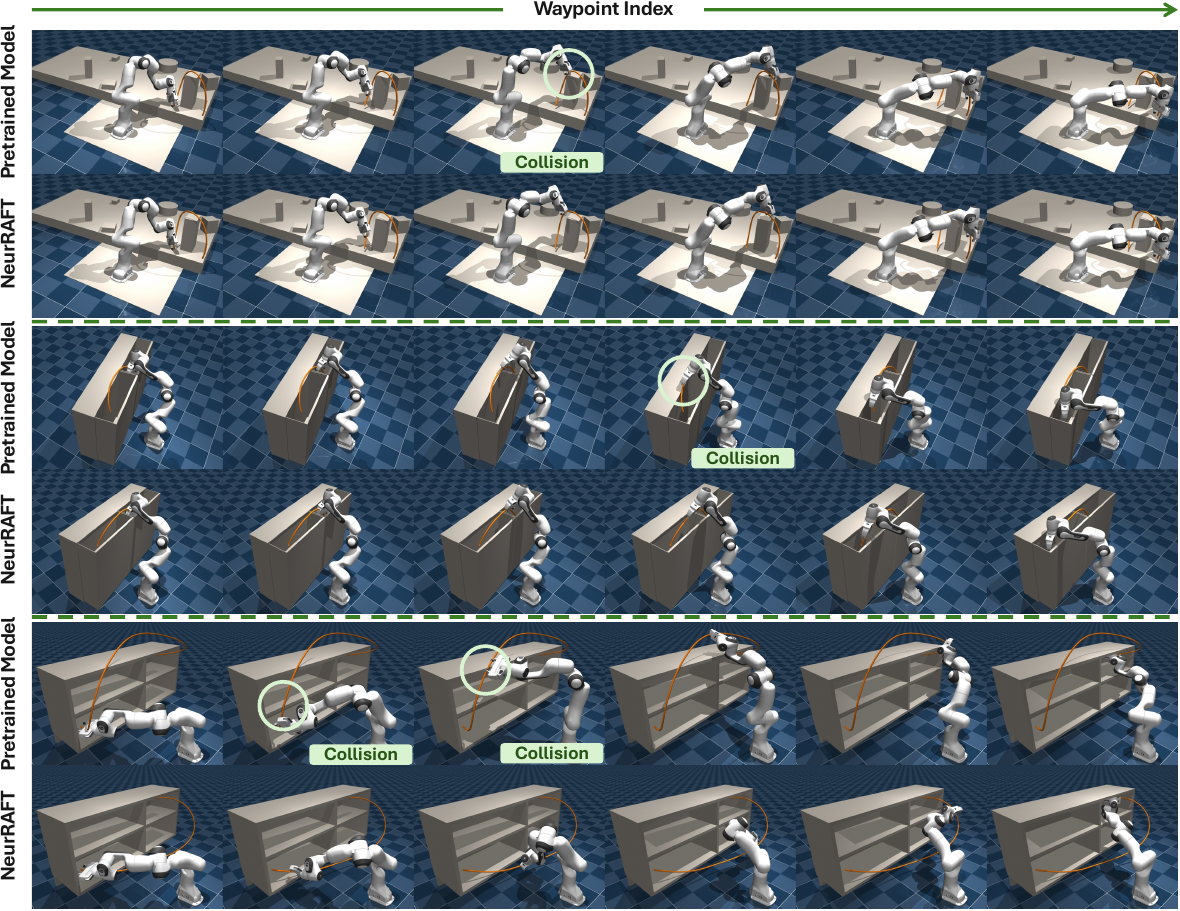}
    \vspace{-0.1in}
    \caption{Qualitative comparison of trajectories generated by the pretrained and preference-aligned models on three representative planning cases. The orange curves denote the TCP trajectories. Across all three cases, preference alignment shifts the trajectory distribution toward higher-clearance solutions, leading the final planner to generate safer trajectories with larger obstacle clearance without any post-hoc trajectory refinement.}
    \label{fig:vis}
    \end{center}
    \vspace{-0.2in}
\end{figure*}

We first evaluate NeurRAFT on the large-scale motion planning benchmark proposed in \cite{fishman2023motion} and compare it against a set of baseline planners, including both classical planners and state-of-the-art neural planners. The success rates on the three test splits are reported in Table~\ref{tab:success_rate}. Trained on the Global Expert demonstrations, NeurRAFT achieves 93.61\%, 97.89\%, and 97.94\% on the Global-Solvable, Hybrid-Solvable, and Both-Solvable splits, respectively. With the Hybrid Expert demonstrations, it achieves 89.56\%, 98.22\%, and 98.17\% on the corresponding splits.

Compared with classical planners, NeurRAFT provides a substantial improvement in success rate. Across the three test splits, classical planners achieve success rates ranging from 26.67\% to 50.22\% on the Global-Solvable split, 31.61\% to 78.44\% on the Hybrid-Solvable split, and 32.20\% to 76.00\% on the Both-Solvable split, whereas NeurRAFT exceeds 89\% in every setting. This performance gap is particularly notable given the difference in the information available at inference time. Classical planners have access to a privileged geometric model of the scene and can perform collision checking during planning, while NeurRAFT directly predicts motion trajectories from a point cloud observation without querying a privileged geometric model.

NeurRAFT also consistently outperforms existing neural motion planners. Among the neural baselines reported in Table~\ref{tab:success_rate}, Neural MP \cite{dalal2024neural} and Cascaded Diffusion \cite{sharma2025cascaded} generally achieve the strongest performance. Moreover, the Global-Solvable split is the most challenging setting, with every planner achieving a lower success rate than on the other two splits. On this split, NeurRAFT achieves 93.61\% with Global Expert training, exceeding Cascaded Diffusion by 8.48 percentage points, and 89.56\% with Hybrid Expert training, exceeding Neural MP by 13.23 percentage points. The margins remain substantial on the other two splits under Global Expert training, where NeurRAFT outperforms Neural MP by 12.39 and 10.27 percentage points on the Hybrid-Solvable and Both-Solvable splits, respectively. Under Hybrid Expert training, the margins become narrower, with improvements of 0.22 percentage points over Cascaded Diffusion on the Hybrid-Solvable split and 1.39 percentage points over Neural MP on the Both-Solvable split, where the strongest baselines already achieve success rates above 96\%.

Table~\ref{tab:success_rate} also isolates the contribution of preference tuning by reporting the performance of the pretrained planner before preference tuning. Under Global Expert training, the pretrained model already surpasses the strongest published baseline by 4.43, 7.67, and 6.66 percentage points on the Global-Solvable, Hybrid-Solvable, and Both-Solvable splits, respectively. Under Hybrid Expert training, it exceeds the strongest baseline by 8.84 percentage points on the Global-Solvable split, while trailing the strongest baselines by only 1.17 and 0.11 percentage points on the Hybrid-Solvable and Both-Solvable splits, respectively. Preference tuning further improves the performance in all six settings, adding 4.05 and 4.39 percentage points on the Global-Solvable split under Global and Hybrid Expert training, respectively, 4.72 and 1.39 percentage points on the Hybrid-Solvable split, and 3.61 and 1.50 percentage points on the Both-Solvable split. Notably, the two settings in which pretraining alone falls short of the strongest baseline both surpass the baseline after preference tuning, while no test setting experiences a performance degradation. A qualitative comparison of trajectories generated by the pretrained and preference-aligned models can be seen in Fig.~\ref{fig:vis}.

Overall, these results support the two-stage design of NeurRAFT. The pretrained planner already matches or exceeds the strongest published baselines while generating a compact set of on-trajectory anchors in a single shot, demonstrating the effectiveness of our neural planner and suggesting that a dense waypoint sequence is not necessary to capture the expert trajectory distribution. Preference tuning further improves performance across all settings, indicating that clearance-based supervision addresses failure modes that remain after imitation learning from positive demonstrations alone. It shapes the output distribution directly, without adding any module or operation at inference.

\subsection{Effect of Candidate Diversity}

\begin{figure}
    \begin{center}
        \includegraphics[width=1.0\columnwidth]{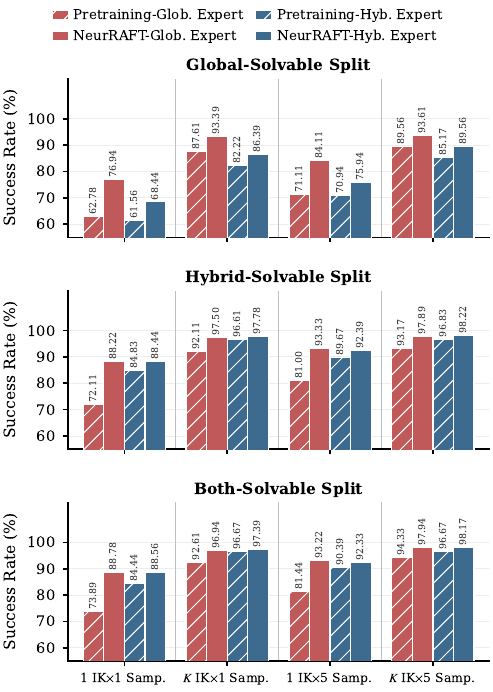}
    \vspace{-0.1in}
    \caption{Success rate before and after clearance-aware preference tuning, across the three test splits and the four candidate-generation settings. Hatched bars report the pretrained planner and solid bars the planner after preference tuning. $K$ denotes the number of feasible inverse-kinematics solutions, and the sample count is the number of trajectories drawn per goal.}
    \label{fig:candidate}
    \end{center}
    \vspace{-0.1in}
\end{figure}

We further evaluate the effect of different candidate-generation strategies, as shown in Fig.~\ref{fig:candidate}. By varying the number of IK solutions and the number of trajectories sampled per IK solution, we consider four settings, namely one sample from a single goal configuration, one sample from each feasible goal configuration, five samples from a single goal, and five samples from each feasible goal configuration. Enlarging the candidate set improves the success rate in every case, and the improvement is larger when the additional candidates come from multiple IK solutions rather than from repeated sampling toward one goal. Under Global Expert training on the Global-Solvable split, the success rate after preference tuning rises from 76.94\% with a single candidate to 84.11\% with five samples toward the same goal, but to 93.39\% with one sample toward each feasible goal. The Hybrid Expert on the Hybrid-Solvable split follows the same ordering, at 88.44\%, 92.39\%, and 97.78\% for the three settings. Combining both diversity sources gives a slightly higher success rate. The ordering holds in all six expert and split combinations and for both the pretrained and the preference-tuned model, indicating that exploiting the kinematic redundancy of the manipulator matters more for planning success than drawing repeated samples toward a fixed goal.

\subsection{Ablation Studies}

\begin{figure}
    \begin{center}
        \includegraphics[width=1.0\columnwidth]{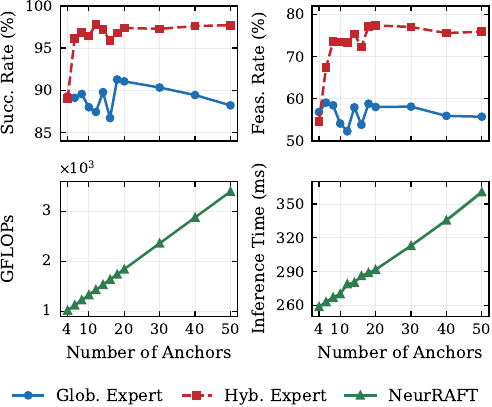}
    \vspace{-0.1in}
    \caption{Effect of the number of anchors on planning quality and inference cost. The top row reports success rate and feasibility rate for models trained on the Global and Hybrid demonstrations, each evaluated on its matching test split. The bottom row reports the computational cost per query, which depends only on the architecture and is therefore shared by both expert settings.}
    \label{fig:waypoint}
    \end{center}
    \vspace{-0.2in}
\end{figure}

We first evaluate the effect of the number of anchors, as shown in Fig.~\ref{fig:waypoint}. We vary $N$ from 4 to 50, where $N=50$ recovers the dense waypoint sequence used in the expert demonstrations, and retrain the model from scratch for each setting. Planning quality from the pretraining stage is largely insensitive to the anchor budget. Except for the smallest setting, where four anchors are too coarse to adequately represent the expert trajectories, both success and feasibility rates remain within a few percentage points across the entire range for both expert settings. In contrast, computational cost increases substantially with the anchor budget. The floating-point cost grows linearly with $N$, accompanied by a corresponding increase in measured inference time. Thus, increasing the number of anchors incurs additional computational cost without providing a consistent improvement in planning quality. We therefore adopt $N=8$ for all remaining experiments, which provides sufficient representation capacity while keeping the computational cost near the lower end of the range.

\begin{table}[htbp]
\caption{Effect of the Jacobian-weighted velocity loss at pretraining under different candidate diversity levels}
\centering
\resizebox{1.0\columnwidth}{!}{%
\begin{threeparttable}

\begin{tabular}{llc|cc|cc|cc}
\toprule

\multicolumn{9}{c}{\textbf{Global Expert}} \\
\midrule

\multirow{2}{*}{\raisebox{-1.0ex}{\textbf{Test}}}
&
\multirow{2}{*}{\raisebox{-1.0ex}{\textbf{Method}}}
&
\textbf{1 IK$\times$1 Samp.}
&
\multicolumn{2}{c|}{\textbf{$K$ IK$\times$1 Samp.}}
&
\multicolumn{2}{c|}{\textbf{1 IK$\times$5 Samp.}}
&
\multicolumn{2}{c}{\textbf{$K$ IK$\times$5 Samp.}}
\\

\cmidrule(lr){3-3}
\cmidrule(lr){4-5}
\cmidrule(lr){6-7}
\cmidrule(lr){8-9}

&
&
\textbf{Succ. / Feas.}
&
\textbf{Succ.} & \textbf{Feas.}
&
\textbf{Succ.} & \textbf{Feas.}
&
\textbf{Succ.} & \textbf{Feas.}
\\

\midrule

\multirow{2}{*}{Glob.}
& w/o Jac.
& 62.11
& 87.00 & 57.24
& \textbf{71.44} & 62.04
& 88.83 & 56.52
\\

& w/ Jac.
& \textbf{62.78}
& \textbf{87.61} & \textbf{58.50}
& 71.11 & \textbf{62.60}
& \textbf{89.56} & \textbf{58.45}
\\

\cmidrule(lr){1-9}

\multirow{2}{*}{Hyb.}
& w/o Jac.
& 70.06
& 90.72 & 63.13
& 78.72 & 69.66
& 92.39 & 62.93
\\

& w/ Jac.
& \textbf{72.11}
& \textbf{92.11} & \textbf{64.82}
& \textbf{81.00} & \textbf{71.56}
& \textbf{93.17} & \textbf{64.33}
\\

\cmidrule(lr){1-9}

\multirow{2}{*}{Both}
& w/o Jac.
& 72.39
& 91.78 & 64.10
& 81.11 & 72.31
& 93.78 & 64.44
\\

& w/ Jac.
& \textbf{73.89}
& \textbf{92.61} & \textbf{65.75}
& \textbf{81.44} & \textbf{73.14}
& \textbf{94.33} & \textbf{65.76}
\\

\midrule
\midrule

\multicolumn{9}{c}{\textbf{Hybrid Expert}} \\
\midrule

\multirow{2}{*}{\raisebox{-1.0ex}{\textbf{Test}}}
&
\multirow{2}{*}{\raisebox{-1.0ex}{\textbf{Method}}}
&
\textbf{1 IK$\times$1 Samp.}
&
\multicolumn{2}{c|}{\textbf{$K$ IK$\times$1 Samp.}}
&
\multicolumn{2}{c|}{\textbf{1 IK$\times$5 Samp.}}
&
\multicolumn{2}{c}{\textbf{$K$ IK$\times$5 Samp.}}
\\

\cmidrule(lr){3-3}
\cmidrule(lr){4-5}
\cmidrule(lr){6-7}
\cmidrule(lr){8-9}

&
&
\textbf{Succ. / Feas.}
&
\textbf{Succ.} & \textbf{Feas.}
&
\textbf{Succ.} & \textbf{Feas.}
&
\textbf{Succ.} & \textbf{Feas.}
\\

\midrule

\multirow{2}{*}{Glob.}
& w/o Jac.
& 60.72
& 82.17 & 52.65
& 70.11 & 61.42
& \textbf{85.67} & 52.79
\\

& w/ Jac.
& \textbf{61.56}
& \textbf{82.22} & \textbf{54.88}
& \textbf{70.94} & \textbf{62.43}
& 85.17 & \textbf{54.90}
\\

\cmidrule(lr){1-9}

\multirow{2}{*}{Hyb.}
& w/o Jac.
& 83.72
& 96.39 & 71.46
& 89.39 & 83.99
& 96.78 & 71.22
\\

& w/ Jac.
& \textbf{84.83}
& \textbf{96.61} & \textbf{73.76}
& \textbf{89.67} & \textbf{84.56}
& \textbf{96.83} & \textbf{73.54}
\\

\cmidrule(lr){1-9}

\multirow{2}{*}{Both}
& w/o Jac.
& \textbf{84.56}
& 96.39 & 72.24
& 89.72 & 84.42
& \textbf{97.33} & 72.24
\\

& w/ Jac.
& 84.44
& \textbf{96.67} & \textbf{74.31}
& \textbf{90.39} & \textbf{85.06}
& 96.67 & \textbf{74.45}
\\

\bottomrule

\end{tabular}

\begin{tablenotes}
\item[*] All values are reported in \%. Succ. and Feas. denote success rate and feasibility rate, respectively. $K$ denotes all feasible IK solutions, Samp. denotes the number of trajectories sampled per goal configuration. Jac. denotes the Jacobian-weighted velocity loss. For the 1 IK $\times$ 1 sample case, only one candidate is generated per problem, so the success rate equals the feasibility rate. Bold values indicate the better result within each pair.
\end{tablenotes}

\label{tab:jacobian_ablation}
\end{threeparttable}
}
\vspace{-0.1in}
\end{table}

\begin{table*}[htbp]
\caption{Success Rate (\%) and Feasibility Rate (\%) of the Global and Hybrid Experts on the Global-Solvable and Hybrid-Solvable test splits, respectively, under different Preference Strengths $\beta$}
\centering
\resizebox{2.0\columnwidth}{!}{%
\begin{threeparttable}
\begin{tabular}{c|c|cc|cc|cc|c|cc|cc|cc}
\toprule

\multirow{3}{*}{\raisebox{-2.0ex}{\textbf{Pref. $\beta$}}}
&
\multicolumn{7}{c|}{\textbf{Global Expert -- Global-Solvable Split}}
&
\multicolumn{7}{c}{\textbf{Hybrid Expert -- Hybrid-Solvable Split}}
\\

\cmidrule(lr){2-8}
\cmidrule(lr){9-15}

&
\textbf{1 IK$\times$1 Samp.}
&
\multicolumn{2}{c|}{\textbf{$K$ IK$\times$1 Samp.}}
&
\multicolumn{2}{c|}{\textbf{1 IK$\times$5 Samp.}}
&
\multicolumn{2}{c|}{\textbf{$K$ IK$\times$5 Samp.}}
&
\textbf{1 IK$\times$1 Samp.}
&
\multicolumn{2}{c|}{\textbf{$K$ IK$\times$1 Samp.}}
&
\multicolumn{2}{c|}{\textbf{1 IK$\times$5 Samp.}}
&
\multicolumn{2}{c}{\textbf{$K$ IK$\times$5 Samp.}}
\\

\cmidrule(lr){2-2}
\cmidrule(lr){3-4}
\cmidrule(lr){5-6}
\cmidrule(lr){7-8}
\cmidrule(lr){9-9}
\cmidrule(lr){10-11}
\cmidrule(lr){12-13}
\cmidrule(lr){14-15}

&
\textbf{Succ. / Feas.}
&
\textbf{Succ.} & \textbf{Feas.}
&
\textbf{Succ.} & \textbf{Feas.}
&
\textbf{Succ.} & \textbf{Feas.}
&
\textbf{Succ. / Feas.}
&
\textbf{Succ.} & \textbf{Feas.}
&
\textbf{Succ.} & \textbf{Feas.}
&
\textbf{Succ.} & \textbf{Feas.}
\\

\midrule

0
&
62.50 & 85.89 & 58.38 & 72.67 & 63.59 & 89.44 & 58.59
&
84.06 & 97.00 & 73.67 & 89.83 & 84.61 & 97.28 & 73.20
\\

1
&
65.22 & 87.56 & 61.87 & 75.22 & 65.69 & 90.56 & 62.09
&
86.50 & 96.83 & 74.83 & 91.44 & 85.75 & 97.33 & 74.46
\\

20
&
73.39 & 92.39 & 75.18 & 82.67 & 73.55 & 93.56 & 75.43
&
89.83 & 98.00 & 78.95 & 92.50 & 88.96 & 97.94 & 78.98
\\

40
&
74.83 & 93.44 & 76.82 & 83.78 & 75.37 & 94.06 & 76.80
&
89.17 & 98.22 & 80.13 & 92.83 & 89.37 & 98.50 & 80.25
\\

60
&
75.44 & 93.00 & 77.37 & 84.67 & 76.58 & 93.83 & 77.20
&
89.61 & 98.00 & 80.51 & 92.67 & 89.55 & 98.17 & 80.54
\\

80
&
75.17 & 93.06 & 77.42 & 84.22 & 76.79 & 94.28 & 77.16
&
88.56 & 97.67 & 80.22 & 92.00 & 89.12 & 98.11 & 80.47
\\

100
&
76.94 & 93.39 & 77.26 & 84.11 & 76.70 & 93.61 & 76.89
&
88.44 & 97.78 & 80.36 & 92.39 & 89.41 & 98.22 & 80.31
\\

\bottomrule
\end{tabular}

\begin{tablenotes}
\item[*] Succ. and Feas. denote the success rate and feasibility rate, respectively. Samp. denotes the number of trajectories sampled per goal configuration, and $K$ denotes the number of feasible solutions returned by inverse kinematics. Pref. denotes the preference strength $\beta$ used during preference tuning. For the 1 IK $\times$ 1 sample case, only one candidate is generated per problem, so the success rate equals the feasibility rate.
\end{tablenotes}

\label{tab:beta_sweep_combined}
\end{threeparttable}
}
\vspace{-0.1in}
\end{table*}

We then evaluate the effectiveness of the Jacobian-weighted velocity loss and report the pretraining results in Table~\ref{tab:jacobian_ablation}. Both variants use the same backbone and training schedule and differ only in whether the kinematics-aware term is included. Across all cases, the Jacobian-weighted term improves both the success and feasibility rates in the vast majority of comparisons, increasing the feasibility rate in 23 of 24 cases and the success rate in 20 of 24 cases, with gains of up to 2.32 and 2.28 percentage points, respectively. In the remaining cases, the performance difference is at most 0.66 percentage points. These results demonstrate that the Jacobian-weighted term consistently improves the quality of the trajectory distribution learned during pretraining.

\begin{figure*}
    \begin{center}
        \includegraphics[width=2.0\columnwidth]{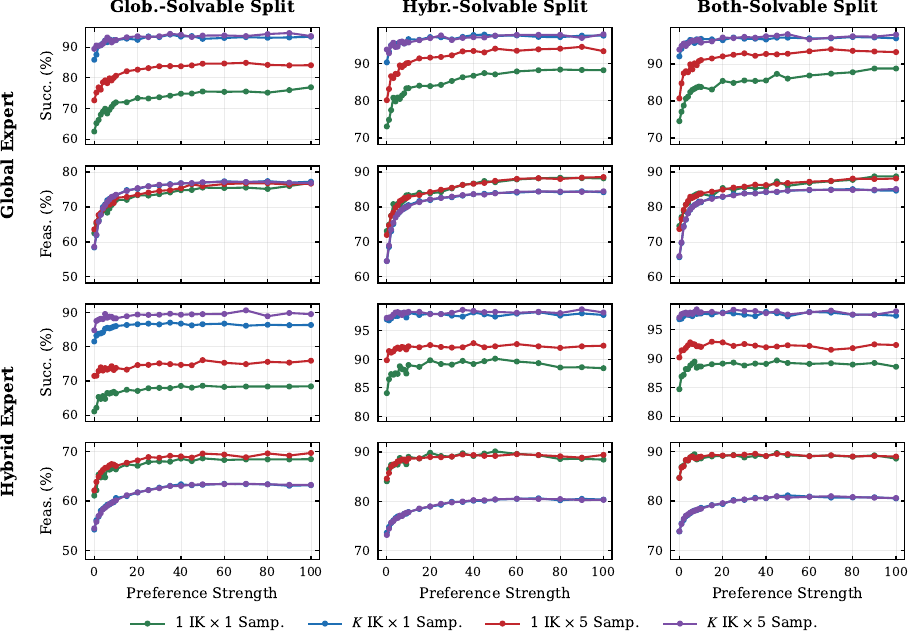}
    \vspace{-0.1in}
    \caption{Success rate and feasibility rate as a function of the preference strength $\beta$. The top two rows report models trained on the Global expert dataset and the bottom two rows models trained on the Hybrid expert dataset, each evaluated on the three test splits of the benchmark. Every panel shows the four candidate-generation settings, with $K$ denoting feasible IK solutions and the sample count denoting trajectories per goal configuration.}
    \label{fig:beta}
    \end{center}
    \vspace{-0.2in}
\end{figure*}

Finally, we examine the effect of the preference strength $\beta$ during the fine-tuning stage and report the results in Table~\ref{tab:beta_sweep_combined} and Fig.~\ref{fig:beta}. We sweep $\beta$ from 0 to 100, where $\beta = 0$ serves as a continued-SFT baseline that indicates whether further training with the standard imitation objective alone provides additional benefit. As shown in Table~\ref{tab:beta_sweep_combined}, continued training at $\beta = 0$ leaves performance essentially unchanged from that of the pretrained planner with the Jacobian-weighted loss in Table~\ref{tab:jacobian_ablation}. Across the four candidate-generation settings and both metrics, the two models differ by 0.02 percentage points on average, with no consistent direction and a largest deviation of 1.72 points. The pretrained model has therefore converged under the imitation objective, and further improvement requires a different training signal. 

When $\beta$ increases from 0 to 100, all metrics improve and gradually converge under both expert settings and across all four candidate-generation configurations. In the single-goal settings, where only one IK solution is used, the gains are large and the two metrics move together. For the Global Expert on the Global-Solvable split, drawing a single sample improves both metrics by 14.44 percentage points, and drawing five samples improves the success rate by 11.44 points and the feasibility rate by 13.11 points. The corresponding gains for the Hybrid Expert on the Hybrid-Solvable split are 4.38 points for a single sample, and 2.56 and 4.80 points for five samples. The two metrics separate once multiple IK solutions are used. For the Global Expert, the feasibility rate then improves by 18.88 and 18.30 points while the success rate improves by only 7.50 and 4.17 points, and the same pattern holds for the Hybrid Expert at 6.69 and 7.11 points against 0.78 and 0.94 points. This difference follows from what the two metrics measure. The clearance-based objective directly improves the quality of individual trajectory samples, which the feasibility rate reports, whereas the success rate depends only on whether the selected candidate is feasible and therefore saturates once the candidate set is diverse enough to already contain a solution. Fig.~\ref{fig:beta} shows the complete sweep, with both metrics increasing rapidly at small $\beta$ and flattening as $\beta$ grows.

\begin{table}[htbp]
\caption{Success rates (\%) on the real-world planning tasks}
\centering
\resizebox{\columnwidth}{!}{%
\begin{threeparttable}
\begin{tabular}{llcccc}
\toprule
\textbf{Expert} & \textbf{Model} & \textbf{Table Top} & \textbf{Drawer} & \textbf{Shelf} & \textbf{Avg.} \\
\midrule
\multirow{3}{*}[-0.4ex]{\textit{Global}}
& Pretraining only        & 85.71 & 82.88 & 70.37 & 79.65 \\
& NeurRAFT                & \textbf{91.43} & \textbf{87.67} & \textbf{79.01} & \textbf{86.04} \\
\cmidrule(lr){2-6}
& \textit{Abs. improv.}   & \textit{+5.72} & \textit{+4.79} & \textit{+8.64} & \textit{+6.39} \\
\midrule
\multirow{3}{*}[-0.4ex]{\textit{Hybrid}}
& Pretraining only        & 90.29 & 78.08 & 53.09 & 73.82 \\
& NeurRAFT                & \textbf{96.57} & \textbf{88.36} & \textbf{75.31} & \textbf{86.75} \\
\cmidrule(lr){2-6}
& \textit{Abs. improv.}   & \textit{+6.28} & \textit{+10.28} & \textit{+22.22} & \textit{+12.93} \\
\bottomrule
\end{tabular}
\begin{tablenotes}
\item[*] All values in \%. Pretraining only denotes the planner before preference tuning, and NeurRAFT denotes the same planner after preference tuning. Abs. improv. is the gain of NeurRAFT over the pretrained planner in percentage points. Avg. is the mean success rate across the three tasks.
\end{tablenotes}
\label{tab:real_robot}
\end{threeparttable}
}
\vspace{-0.1in}
\end{table}

\subsection{Real-World Experiments}

We deploy NeurRAFT on a physical Franka robot manipulator to evaluate whether a planner trained entirely in simulation can transfer to real-world planning problems without retraining or adaptation. We construct three scene categories with increasing environmental complexity and planning difficulty: a tabletop, a drawer, and a shelf. Within each category, we vary obstacle types and placements, as well as the start configurations and target end-effector poses, resulting in more than 500 real-world planning problems in total. The scene is observed using three depth cameras, and the robot points are sampled in the same manner as during training.

Table~\ref{tab:real_robot} reports the success rates among three scenarios. Preference tuning improves performance on every task under both expert settings, with average gains of 6.39 percentage points for the Global Expert and 12.93 points for the Hybrid Expert, showing the same trend observed in the benchmark results. Moreover, NeurRAFT achieves average success rates of 86.04\% and 86.75\% under the Global and Hybrid Expert settings, respectively, demonstrating that the learned trajectory distribution transfers zero-shot to a physical robot despite noisy and partially occluded depth observations. The performance is approximately 10 percentage points lower than in simulation, reflecting the challenges introduced by the real-world deployment setting. Two factors primarily account for this gap. First, the point clouds used during training are clean and complete, whereas real-world observations are captured from fixed viewpoints and contain substantial noise and occlusions between objects. Second, the physical obstacles have shapes and dimensions that differ from the simulated training instances. The planner therefore encounters both a sensing gap and a geometry gap, yet retains most of its simulated performance under both expert settings.

\section{Conclusion}

This paper presented NeurRAFT, a generative motion planner that operates over a compact set of anchor waypoints rather than a dense trajectory sequence. The model generates all anchors in a single shot through flow matching, keeps the start and goal anchors fixed to their given configurations throughout generation, and recovers a smooth trajectory by cubic-spline interpolation. A Jacobian-weighted term in the training objective penalizes joint-space errors in proportion to the end-effector displacement they induce. On top of the pretrained planner, we introduced a clearance-aware preference-tuning stage that ranks the model's own samples by signed obstacle clearance and fine-tunes toward safer paths with direct preference optimization. The preference data are produced by the model itself on the existing training set, so the stage requires no additional demonstrations, no human labels, and no online interaction.

Experiments on the M$\pi$Nets benchmark show that NeurRAFT outperforms both classical and state-of-the-art neural planners on all three test splits and under both expert data pipelines. The pretrained planner alone already exceeds the strongest published baselines in most settings, which indicates that a dense waypoint sequence is not necessary to capture the expert trajectory distribution. Preference tuning improves every setting on top of that, and the ablation on the preference strength shows that increasing preference strength consistently improves feasibility until convergence, demonstrating the effectiveness of preference alignment in promoting feasible trajectories. Real-world experiments on more than 500 planning problems confirm that the planner transfers zero-shot to a physical Franka manipulator under noisy and partially occluded depth observations, and that the ordering between the two training stages is preserved.

Despite its strong performance, several limitations remain. First, NeurRAFT currently assumes a static environment and generates a full trajectory from a fixed observation. When the environment changes during execution, the planner must replan from scratch rather than continuously adapting the trajectory. Future work will investigate reactive planning mechanisms that allow NeurRAFT to adapt its planning result in response to dynamic obstacles. Second, although NeurRAFT demonstrates strong zero-shot transfer to real-world environments, the remaining simulation-to-real gap leaves room for improving robustness to perception noise and unseen geometries. Future work could explore lightweight domain randomization or post-training with real-world preference data to further narrow this gap.

\bibliographystyle{IEEEtran}
\bibliography{ref}{}

\end{document}